\documentclass{article} 
\usepackage[OT1]{fontenc}
\usepackage{iclr2027_preprint,times}

\usepackage{amsmath,amssymb}
\IfFileExists{math_commands.tex}{

\usepackage{amsmath,amsfonts,bm}

\def\eqref#1{equation~\ref{#1}}

\def\1{\bm{1}}

\DeclareMathAlphabet{\mathsfit}{\encodingdefault}{\sfdefault}{m}{sl}
\SetMathAlphabet{\mathsfit}{bold}{\encodingdefault}{\sfdefault}{bx}{n}

}{}

\usepackage{hyperref}
\hypersetup{hidelinks}
\usepackage{url}
\usepackage{booktabs,graphicx}
\usepackage{tikz}

\title{When Text Matters: Design Principles for Visual\\ Token Pruning in Vision-Language Models}

\author{
\textbf{Minchan Kang}\textsuperscript{1}, \textbf{Kyeonghye Park}\textsuperscript{1}, \textbf{Seoyoung Cho}\textsuperscript{1}, \textbf{Daeshik Kim}\textsuperscript{1}\thanks{Equal correspondence: Daeshik Kim and Yucheol Cho.}, \textbf{Yucheol Cho}\textsuperscript{2}\footnotemark[1] \\
\textsuperscript{1}Korea Advanced Institute of Science and Technology (KAIST), \textsuperscript{2}Hanbat National University \\
{\small\texttt{\{mc.kang,pkhpjhs,52tjdud,daeshik\}@kaist.ac.kr; yccho@hanbat.ac.kr}}
}

\usetikzlibrary{shapes.geometric}
\begin{document}

\maketitle
\begin{abstract}
Visual token pruning has been widely studied as a practical approach to reducing the computational cost of large vision-language models. However, it struggles to preserve essential visual information, which can lead to substantial performance degradation. In particular, image-based token selection can overlook task-relevant details, while text-guided token selection may fail to capture the text--visual relationships needed for complex reasoning. We find that applying textual guidance too early can limit its ability to identify answer-relevant visual regions, whereas text-to-visual attention becomes more informative at intermediate decoder depths. This finding motivates our training-free method, which separates early vision-guided pruning from deferred text-guided reselection. We first prune visual tokens using vision-encoder attention, retain additional candidates until the decoder midpoint, and then use text-to-visual attention to determine the final visual-token set. Across eight benchmarks and three models, our method outperforms the best-performing baselines by an average of 11.10 and 16.84 percentage points in performance recovery at 80\% and 90\% pruning, respectively, with comparable or lower LLM-prefill latency than most baselines.
The source code is publicly available at \url{https://github.com/kmc3661/DeFT}.

\end{abstract}

\section{Introduction}

Large vision-language models (VLMs) support diverse tasks, including visual question answering, image captioning, and multimodal reasoning \citep{liu2023visual,dai2023instructblip}. Their capabilities, however, come with substantial computational costs: representing images with numerous visual tokens increases the sequence length processed by the language model, particularly for high-resolution inputs \citep{chen2024image,wang2024qwen2}. Visual token pruning addresses this overhead by shortening the visual sequence. Training-free approaches are especially practical, as they reduce inference costs without updating model parameters or training an additional selector \citep{chen2024image,zhang2024sparsevlm}.

The central challenge is deciding which visual information to preserve. Image-based token selection uses visual importance, redundancy, or sensitivity to identify tokens without conditioning on the prompt \citep{yang2025visionzip,kim2026zoo}. Text-guided token selection instead uses interactions between textual and visual tokens to identify information relevant to the requested task \citep{xing2024pyramiddrop,zhang2024sparsevlm}. Despite these differences, both approaches can discard essential visual information, leading to substantial performance degradation.

\begin{figure}[t]
\centering
\includegraphics[width=\linewidth]{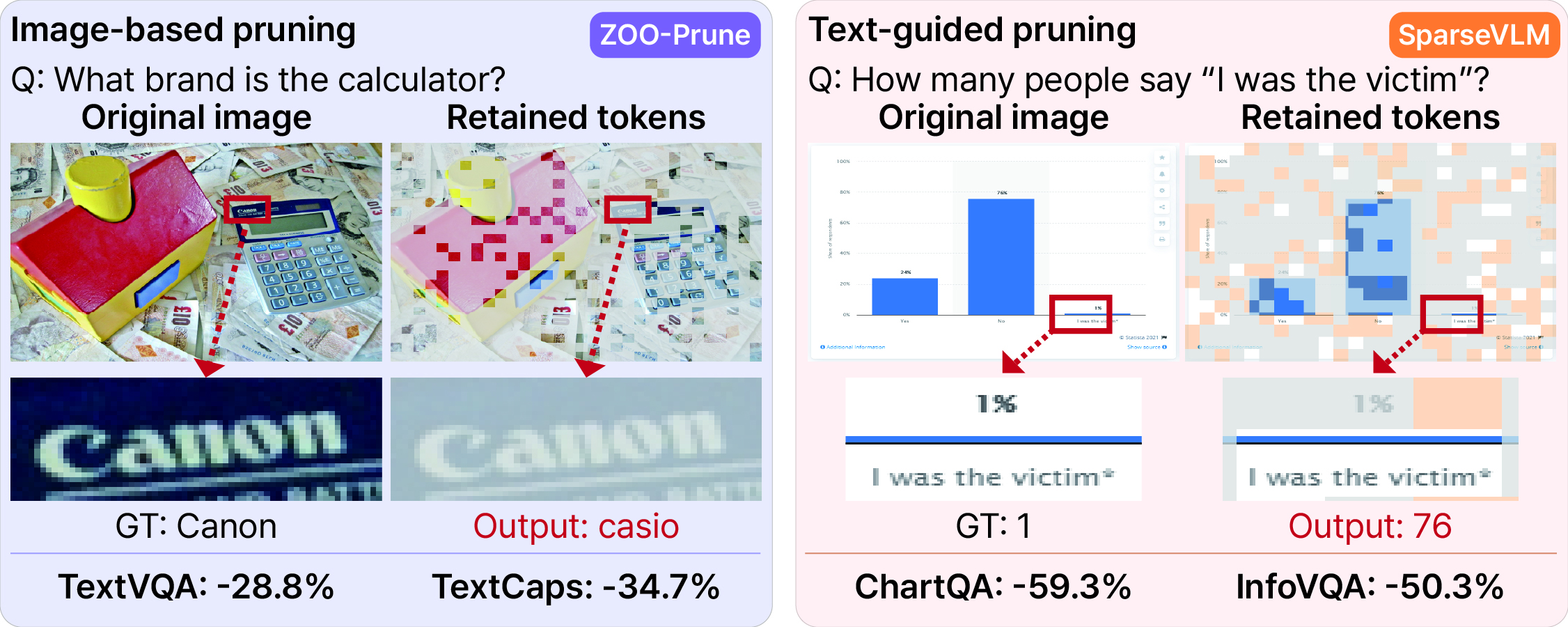}
\caption{
Failure cases on Qwen3-VL-8B at 80\% pruning.
Image-based visual token pruning discards the visual region containing the requested text, while text-guided visual token pruning preserves related support but loses the corresponding value.
Gray: discarded regions; orange: merged support.
Percentages denote full-benchmark performance losses relative to Dense.
}
\label{fig:motivation}
\end{figure}

Figure~\ref{fig:motivation} illustrates representative failure cases of image-based and text-guided visual token pruning. For image-based pruning, ZOO-Prune \citep{kim2026zoo} retains prominent visual features of the calculator but discards the region containing the brand name requested by the question. For text-guided pruning, SparseVLM \citep{zhang2024sparsevlm} retains support for the phrase ``I was the victim'' but loses the corresponding value, illustrating a difficulty in capturing the relationships between textual and visual information needed to answer the question. Full-benchmark evaluations further reveal substantial performance losses. At 80\% pruning on Qwen3-VL-8B, ZOO-Prune loses 28.8\% and 34.7\% relative to Dense on TextVQA and TextCaps, respectively, which rely on scene-text understanding \citep{textvqa,textcaps}. SparseVLM loses more than 50\% on ChartQA and InfoVQA, which require complex reasoning over relationships among textual and visual elements \citep{chartqa,infovqa}.

These failures highlight the difficulty of identifying the visual information a VLM needs to reason toward a correct answer. This raises a key question: how can textual guidance be used more effectively to identify visual evidence that is relevant to the prompt? To investigate this, we analyze the ability of text-to-visual attention to distinguish answer-relevant regions across decoder depths. We find that this ability becomes substantially stronger at intermediate layers than near the input (Section~\ref{sec:depth-rationale}), indicating that the timing of textual guidance is critical to effective token selection.

Building on this finding, we introduce a simple training-free method that combines early image-based pruning with deferred text-guided selection. Specifically, we first prune visual tokens before the LLM using vision-encoder attention, while retaining additional candidates beyond the final token budget. At the decoder midpoint, text-to-visual attention reselects from these candidates to form the final token set. This design deliberately defers textual guidance to the midpoint until it becomes more informative, while relying solely on image information for early pruning. Across eight benchmarks and three VLMs, our method consistently achieves the highest average Dense-relative performance retention across all evaluated pruning ratios, with the margin over the strongest baseline increasing to 11.10 and 16.84 percentage points at 80\% and 90\% pruning, respectively.

\section{Related Work}
\subsection{Large Vision-Language Models}
VLMs extend large language models \citep{brown2020language,grattafiori2024llama} with visual encoders and multimodal alignment \citep{li2023blip}. CogVLM introduces visual expert modules for deeper vision--language fusion, while Idefics2 studies architecture and training choices for effective multimodal learning \citep{wang2024cogvlm,laurenccon2024matters}. Models such as LLaVA-1.5, LLaVA-OneVision, and InternVL advance visual understanding through improved architectures, training, and visual representations \citep{liu2024improved,li2024llava,chen2024internvl}, while high-resolution processing increases visual-token costs \citep{guo2024llava}. We evaluate our token pruning method on Qwen3-VL and LLaVA-OneVision-1.5 \citep{qwen3vl,onevision15}.

\subsection{Visual Token Pruning}
Reducing the computational cost of VLMs has been explored through knowledge distillation \citep{hinton2015distilling,cao2025move}, quantization \citep{li2025mbq,xiang2026fine}, and visual token reduction. Token reduction encompasses merging in vision transformers \citep{bolya2022token,norouzi2024algm} and training-based pruning and merging for VLMs \citep{cao2023pumer,cao2024madtp}. Training-free approaches use pretrained signals to select tokens, drawing on visual importance, diversity, and sensitivity \citep{zhang2025beyond}, text--visual attention and instruction-conditioned relevance \citep{zhang2026beyond,liang2026pyramid}, or combinations of visual saliency, textual relevance, and diversity \citep{liu2026crisprune}. Complementary strategies preserve information through merging or recycling \citep{yang2025vflowopt} and correct token-reduction distortions \citep{cho2026restore}. Building on these studies, we examine when textual guidance becomes informative for visual token selection, motivating a separation between early visual pruning and later text-guided reselection.

\section{Method}
\label{sec:method}

\begin{figure}[t]
\centering
\includegraphics[width=\linewidth]{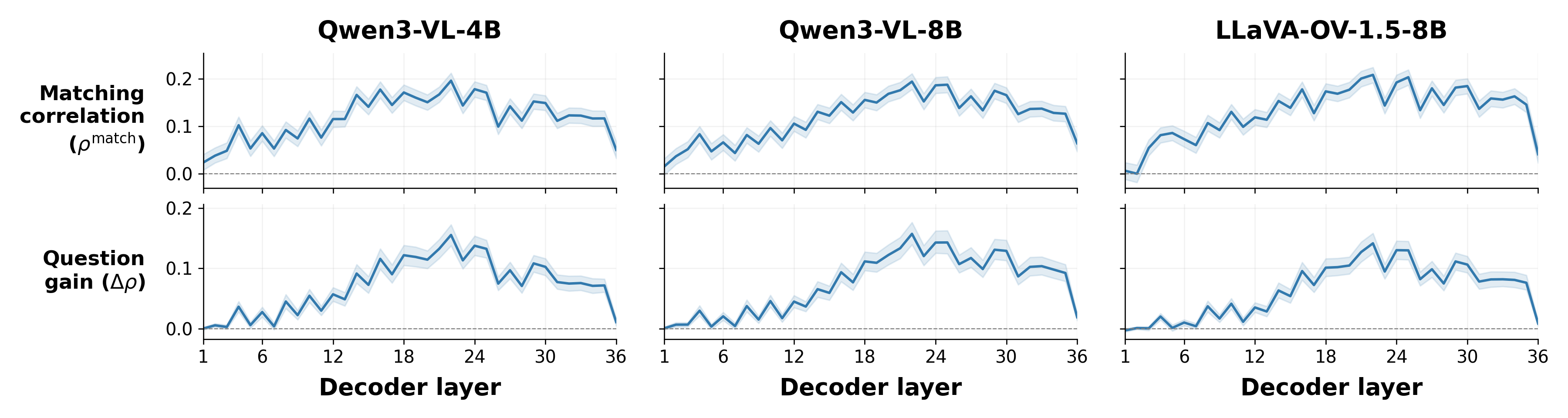}
\caption{
Text-guided evidence identification across decoder depth. Top: correlation with masking-based region importance. Bottom: matching-versus-swapped question gain. Results average four QA benchmarks equally, each with 300 images and two questions per image.
}
\label{fig:question-depth-design}
\end{figure}

\begin{table}
\centering
\small
\caption{
Selection-depth ablation at 80\% pruning under matched decoder visual-token processing budgets.
Observed peak denotes block 22, where the mean question gain in Figure~\ref{fig:question-depth-design} is highest.
Values are eight-task Dense-relative recovery (\%), with the best results in bold.
}
\label{tab:layer-ablation}
\begin{tabular}{@{}lrrrr@{}}
\toprule
Boundary & Qwen3-VL-4B & Qwen3-VL-8B & LLaVA-OV-1.5-8B & Mean \\
\midrule
$D/4$ (9 blocks) & 82.119 & 86.551 & 85.898 & 84.856 \\
$2D/4$ (18 blocks) & \textbf{90.860} & \textbf{94.787} & \textbf{92.769} & \textbf{92.806} \\
Observed peak (22 blocks) & 89.185 & 94.014 & 92.299 & 91.833 \\
$3D/4$ (27 blocks) & 87.508 & 93.431 & 91.162 & 90.701 \\
\bottomrule
\end{tabular}
\end{table}

\subsection{Motivation and Design Rationale}
\label{sec:depth-rationale}

To determine when text-to-visual attention provides a reliable signal for token selection, we evaluate its ability to identify answer-relevant visual regions across decoder depth on four multimodal QA benchmarks \citep{textvqa,chartqa,infovqa,ai2d}. For each benchmark, we sample 300 images with two distinct questions per image and divide each image's native visual-token grid into eight spatial regions. Using the unpruned model (Dense), we mask one region at a time throughout all decoder blocks and measure its effect on the correct answer. Let $\mathcal L_{\mathrm{NLL}}(\mathcal M)$ denote the token-averaged negative log-likelihood of the correct answer when regions $\mathcal M$ are masked. We define the importance of region $\mathcal R_r$ as

\begin{equation}
h_r =
\mathcal L_{\mathrm{NLL}}(\mathcal R_r)
-
\mathcal L_{\mathrm{NLL}}(\varnothing),
\label{eq:region-importance}
\end{equation}

where $\varnothing$ denotes the unmasked condition. A larger $h_r$ means that masking region $\mathcal R_r$ has a larger negative effect on answering the question. We next ask whether text-to-visual attention identifies these important regions and whether this ability is specifically conditioned on the question. For this analysis, we use two different questions $q$ and $q'$ about each image. For both questions, we define
\begin{equation}
h^u=[h_1^u,\ldots,h_8^u],
\qquad
s_\ell^u=[s_{\ell,1}^u,\ldots,s_{\ell,8}^u],
\qquad
u\in\{q,q'\},
\label{eq:region-vectors}
\end{equation}
where $h^u$ contains the masking-based importance of the eight regions with respect to the correct answer for question $u$, and $s_\ell^u$ contains their region-level text-to-visual attention at decoder block $\ell$.

We first test whether each question's attention ranks the important regions for answering that question. Let $\rho(\cdot,\cdot)$ denote Spearman rank correlation. Averaging over the two questions, we define
\begin{equation}
\rho_\ell^{\mathrm{match}}
=
\frac{1}{2}
\left[
\rho(s_\ell^q,h^q)
+
\rho(s_\ell^{q'},h^{q'})
\right].
\label{eq:matching-correlation}
\end{equation}
A larger $\rho_\ell^{\mathrm{match}}$ indicates better agreement between text-to-visual attention and the corresponding region importance.

We next test whether this alignment is specific to the question by comparing each question's region importance with attention from the other question while keeping the image fixed:

\begin{equation}
\rho_\ell^{\mathrm{swap}}
=
\frac{1}{2}
\left[
\rho(s_\ell^{q'},h^q)
+
\rho(s_\ell^q,h^{q'})
\right],
\qquad
\Delta\rho_\ell
=
\rho_\ell^{\mathrm{match}}
-
\rho_\ell^{\mathrm{swap}}.
\label{eq:question-specificity}
\end{equation}
Here, $\rho_\ell^{\mathrm{swap}}$ measures how well attention from the other question ranks the important regions for the current question. A positive $\Delta\rho_\ell$ therefore means that the matching question identifies its own important regions better than the other question. Figure~\ref{fig:question-depth-design} reports $\rho_\ell^{\mathrm{match}}$ in the top row and $\Delta\rho_\ell$ in the bottom row. Both are weak in the early decoder layers but become substantially stronger at intermediate depths. This shows that text-to-visual attention provides limited guidance near the decoder input, while becoming much more effective at identifying question-relevant visual evidence after sufficient decoder processing.

These results motivate deferring text-guided selection beyond the early decoder layers. The reselection depth must also account for efficiency: a deeper boundary processes retained visual tokens through more decoder blocks, so fewer can be retained under the same visual-token processing budget. We therefore ablate the reselection depth in our method under matched budgets. Table~\ref{tab:layer-ablation} shows that the decoder midpoint achieves the highest recovery across all three models. Together, the depth-wise analysis and budget-controlled ablation motivate the decoder midpoint as the reselection boundary, where textual guidance is informative while sufficient candidates can be retained at practical cost. We therefore use $L=D/2$ for text-guided reselection in our method.

\begin{figure}[t]
\centering
\includegraphics[width=\linewidth]{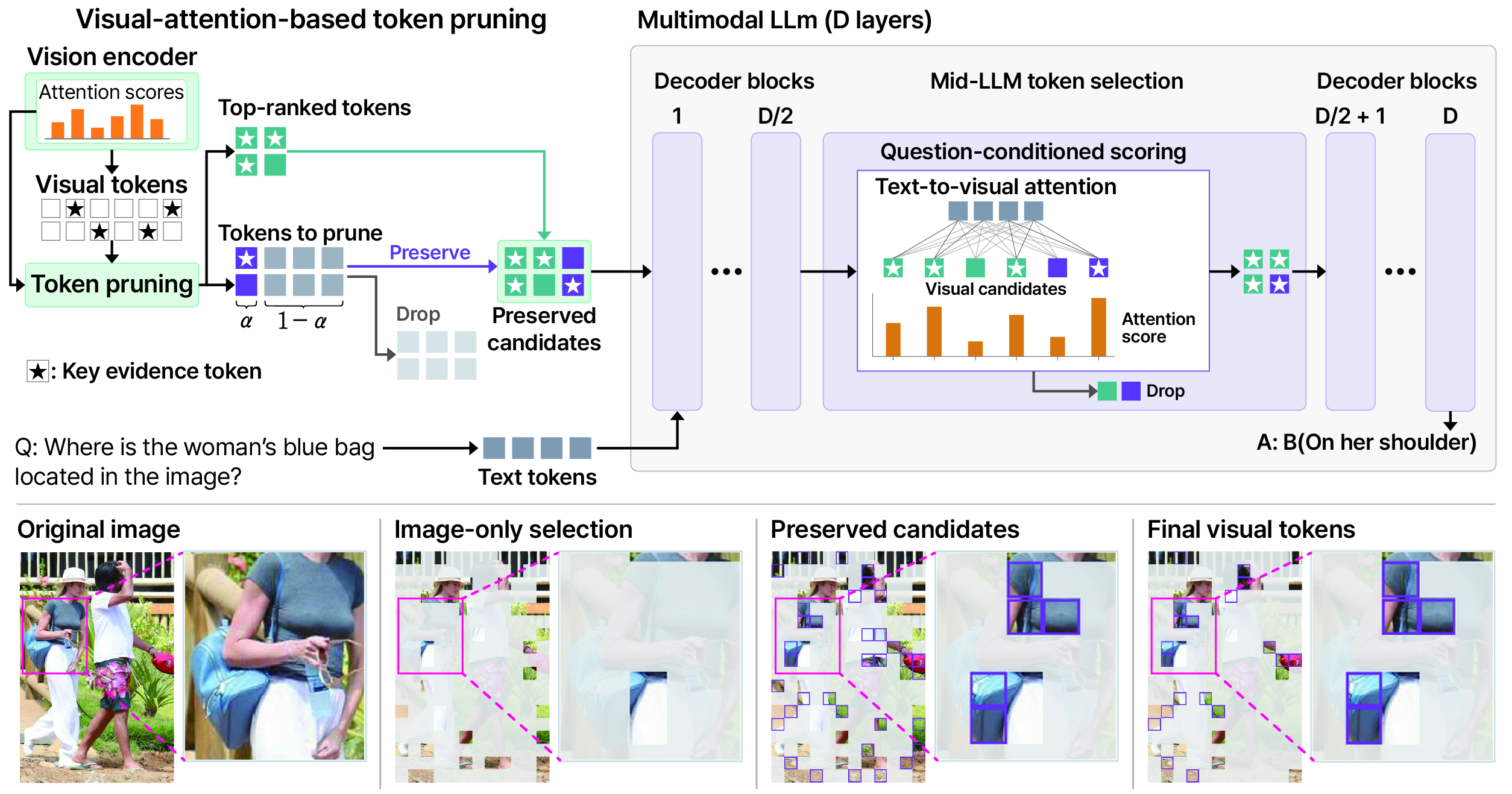}
\caption{
Overview of our two-stage pruning.
Before the LLM, vision-encoder attention guides visual-token pruning while preserving an expanded candidate set.
At the decoder midpoint, text-to-visual attention selects the final visual-token set.
The reserve fraction $\alpha\in[0.1,0.3]$ controls the number of additional candidates; we use $\alpha=0.2$ by default.
}
\label{fig:overall}
\end{figure}

\subsection{Proposed Method}
\label{sec:proposed-method}

\paragraph{Early visual pruning with reserve.}
As illustrated in Figure~\ref{fig:overall}, our method first performs vision-guided pruning before the LLM while preserving additional visual candidates for later reselection. Prior training-free pruning methods have shown that attention within the vision encoder provides an effective signal for identifying visually informative tokens \citep{yang2025visionzip,zhang2025beyond,kim2026zoo}.  We therefore use this signal for pre-LLM pruning, avoiding decoder computation. Unlike methods that incorporate text--visual interactions into early token selection \citep{xing2024pyramiddrop,zhang2024sparsevlm}, we keep this stage purely vision-based, following our observation in Section~\ref{sec:depth-rationale} that textual guidance is less informative in early decoder layers.

Let $N$ be the number of prunable visual tokens, $D$ the decoder depth, and $K\approx(1-p)N$ the final token budget at pruning ratio $p$. We first obtain the head-averaged vision-encoder attention
\begin{equation}
\mathbf A^{\mathrm{img}}
=
\frac{1}{H}\sum_{h=1}^{H}
\operatorname{softmax}
\left(
\frac{\mathbf Q_h\mathbf K_h^{\top}}{\sqrt{d_h}}
\right),
\qquad
s_i^{\mathrm{img}}
=
\begin{cases}
A^{\mathrm{img}}_{\mathrm{cls},i}, & \text{with a class token},\\[2pt]
\sum_{j\in\mathcal V}A^{\mathrm{img}}_{j,i}, & \text{otherwise},
\end{cases}
\label{eq:image-score}
\end{equation}
where $\mathcal V$ denotes the visual-token positions. Thus, class-token-based encoders use CLS-to-visual attention, whereas encoders without a class token use the self-attention received by each visual token from other visual queries.

Rather than immediately pruning to $K$ tokens, we preserve additional image-ranked candidates:
\begin{equation}
M=\min\{N,K+\lceil\alpha(N-K)\rceil\},
\qquad
\mathcal C=\operatorname{TopK}_{M}(s^{\mathrm{img}}),
\label{eq:reserve}
\end{equation}
where $\alpha\in[0.1,0.3]$ is a scalar hyperparameter that controls the fraction of otherwise pruned tokens preserved as additional candidates; we use $\alpha=0.2$ by default. The set $\mathcal C$ contains the image-only top-$K$ together with the next $M-K$ candidates. We process these candidates until the decoder midpoint $L=D/2$, preserving alternatives that may become important once textual guidance becomes sufficiently informative.

\paragraph{Deferred text-guided reselection.}
At the decoder midpoint, we revise the image-only selection with text-to-visual attention. Let $\mathcal T$ and $\mathcal C$ denote the input text tokens and visual candidates, respectively. Using the decoder native queries and keys, we compute the head-averaged attention as
\begin{equation}
\mathbf A^{\mathrm{text}}
=
\frac{1}{H}\sum_{h=1}^{H}
\operatorname{softmax}
\left(
\frac{\mathbf Q^{\mathcal T}_h\mathbf K_h^{\top}}{\sqrt{d_h}}
\right),
\qquad
s_i^{\mathrm{text}}
=
\sum_{t\in\mathcal T}A^{\mathrm{text}}_{t,i},
\label{eq:text-score}
\end{equation}
where the softmax follows the decoder's native attention mask and $s_i^{\mathrm{text}}$ is the attention received by candidate $i$ from the input text tokens. We select the final visual-token set as
\begin{equation}
\mathcal S
=
\operatorname{TopK}_{K}
\left(
s^{\mathrm{text}}\big|_{\mathcal C}
\right).
\label{eq:text-selection}
\end{equation}

Overall, image scores are used only to construct the candidate set $\mathcal C$, while the final selection at the decoder midpoint depends solely on text-to-visual attention. This allows candidates outside the initial vision-guided top-$K$ to enter the final set when they become more relevant to the input. By separating early vision-guided pruning from deferred text-guided reselection, the method reduces visual-token processing early without committing to the final token set before textual guidance becomes sufficiently informative.

\section{Experiments}
\label{sec:experiments}

\subsection{Experimental Setup}

\paragraph{Models and baselines.}
We evaluate Qwen3-VL-4B, Qwen3-VL-8B \citep{qwen3vl}, and LLaVA-OneVision-1.5-8B \citep{onevision15}. We compare with five training-free visual-token reduction methods: FastV \citep{chen2024image}, SparseVLM \citep{zhang2024sparsevlm}, VisPruner \citep{zhang2025beyond}, ZOO-Prune \citep{kim2026zoo}, and RESTORE \citep{cho2026restore}. Since SparseVLM progressively changes the visual-token count across decoder depth, we match its average decoder token usage to ours for a comparable processing budget.

\paragraph{Benchmarks.}
We evaluate eight full benchmarks: TextVQA \citep{textvqa} for scene-text understanding; ChartQA \citep{chartqa} and InfoVQA \citep{infovqa} for chart and infographic reasoning; AI2D \citep{ai2d}, MMMU \citep{mmmu}, and MMStar \citep{mmstar} for diagram and general multimodal reasoning; and NoCaps \citep{nocaps} and TextCaps \citep{textcaps} for open-ended captioning. Together, they test preservation of both global context and localized key evidence.

\paragraph{Implementation and metrics.}
We evaluate 70\%, 80\%, and 90\% visual-token pruning with midpoint reselection. We report each benchmark's native metric, while \emph{Avg. Rel.} is the mean task score normalized by the corresponding Dense score; additional details are in Appendix~\ref{app:protocol}.

\begin{table}[t]
\centering
\caption{Qwen3-VL-4B results across eight benchmarks. Avg. Rel.: mean performance relative to Dense (\%); higher is better. Bold/underline: best/second-best pruned results at each ratio.}
\label{tab:main-qwen3-4b}
\begingroup
\resizebox{\linewidth}{!}{%
\begin{tabular}{@{}lrrrrrrrrr@{}}
\toprule
Method & TextVQA & MMMU & AI2D & MMStar & ChartQA & NoCaps & TextCaps & InfoVQA & Avg. Rel. \\
\midrule
Dense & 81.30 & 46.22 & 81.87 & 56.40 & 82.92 & 66.51 & 81.86 & 77.43 & 100.00 \\
\midrule
\multicolumn{10}{c}{\textit{Prune 70\% of vision tokens}} \\
\midrule
FastV (ECCV2024) & \underline{74.89} & 43.33 & 72.22 & 46.53 & 45.64 & 63.16 & \textbf{84.84} & 44.51 & 83.46 \\
SparseVLM (ICML2025) & 73.21 & \textbf{45.44} & 73.87 & 48.80 & 50.72 & 65.68 & 74.94 & 45.78 & 84.46 \\
VisPruner (ICCV2025) & 64.76 & 44.33 & 72.31 & 47.07 & 65.00 & 66.45 & 68.25 & 46.48 & 83.63 \\
ZOO-Prune (CVPR2026) & 65.64 & \underline{45.11} & \underline{76.75} & \underline{51.20} & \underline{72.84} & 66.12 & 58.97 & \underline{53.92} & \underline{86.48} \\
RESTORE (ICML2026) & 66.17 & 43.89 & 72.83 & 50.67 & 55.44 & \textbf{67.03} & 73.34 & 37.77 & 82.64 \\
\textbf{Ours} & \textbf{77.57} & 44.78 & \textbf{78.79} & \textbf{52.67} & \textbf{75.92} & \underline{66.59} & \underline{81.56} & \textbf{61.32} & \textbf{94.05} \\
\midrule
\multicolumn{10}{c}{\textit{Prune 80\% of vision tokens}} \\
\midrule
FastV (ECCV2024) & \underline{68.57} & 42.44 & 66.84 & 42.00 & 28.72 & 58.98 & \textbf{83.07} & 33.93 & 75.11 \\
SparseVLM (ICML2025) & 66.59 & \underline{44.11} & 71.60 & 46.93 & 34.44 & 63.75 & 67.98 & 38.35 & 77.24 \\
VisPruner (ICCV2025) & 48.16 & 42.56 & 67.58 & 44.93 & 49.08 & 65.61 & 50.76 & 36.53 & 72.57 \\
ZOO-Prune (CVPR2026) & 56.14 & 44.00 & \underline{72.44} & 47.73 & \underline{63.16} & 65.91 & 52.04 & \underline{43.66} & \underline{79.07} \\
RESTORE (ICML2026) & 55.64 & 43.67 & 69.11 & \underline{48.07} & 42.12 & \textbf{66.58} & 63.86 & 30.62 & 75.13 \\
\textbf{Ours} & \textbf{74.91} & \textbf{45.11} & \textbf{76.78} & \textbf{50.53} & \textbf{71.20} & \underline{66.57} & \underline{80.57} & \textbf{53.73} & \textbf{90.86} \\
\midrule
\multicolumn{10}{c}{\textit{Prune 90\% of vision tokens}} \\
\midrule
FastV (ECCV2024) & 46.46 & 39.22 & 63.08 & 33.00 & 15.60 & 47.82 & \underline{70.71} & 25.46 & 60.94 \\
SparseVLM (ICML2025) & \underline{48.95} & \underline{43.78} & \underline{67.52} & 42.07 & 19.36 & 57.96 & 53.31 & 28.15 & 65.49 \\
VisPruner (ICCV2025) & 30.69 & 41.78 & 66.48 & 38.27 & 32.52 & 62.27 & 35.74 & 27.14 & 61.09 \\
ZOO-Prune (CVPR2026) & 40.33 & 42.67 & 67.36 & 43.67 & \underline{44.32} & 63.84 & 41.14 & \underline{30.78} & \underline{67.63} \\
RESTORE (ICML2026) & 37.81 & \underline{43.78} & 65.77 & \underline{43.80} & 28.12 & \textbf{65.08} & 45.93 & 25.71 & 65.04 \\
\textbf{Ours} & \textbf{69.27} & \textbf{44.11} & \textbf{72.96} & \textbf{47.67} & \textbf{55.68} & \underline{63.99} & \textbf{73.21} & \textbf{43.54} & \textbf{82.91} \\
\bottomrule
\end{tabular}%
}
\endgroup
\end{table}

\begin{table}[t]
\centering
\caption{Qwen3-VL-8B results across eight benchmarks. Avg. Rel.: mean performance relative to Dense (\%); higher is better. Bold/underline: best/second-best pruned results at each ratio.}
\label{tab:main-qwen3}
\begingroup
\resizebox{\linewidth}{!}{%
\begin{tabular}{@{}lrrrrrrrrr@{}}
\toprule
Method & TextVQA & MMMU & AI2D & MMStar & ChartQA & NoCaps & TextCaps & InfoVQA & Avg. Rel. \\
\midrule
Dense & 82.95 & 51.67 & 83.68 & 62.73 & 83.48 & 63.54 & 81.62 & 81.19 & 100.00 \\
\midrule
\multicolumn{10}{c}{\textit{Prune 70\% of vision tokens}} \\
\midrule
FastV (ECCV2024) & 74.54 & 49.44 & 72.73 & 50.27 & 43.60 & 63.11 & \textbf{84.87} & 42.53 & 82.56 \\
SparseVLM (ICML2025) & \underline{76.12} & 51.33 & 76.68 & 54.93 & 47.52 & 63.54 & 78.13 & 48.95 & 85.41 \\
VisPruner (ICCV2025) & 75.85 & \textbf{51.56} & \underline{78.47} & 52.87 & 66.68 & 63.13 & 79.61 & 52.60 & \underline{88.85} \\
ZOO-Prune (CVPR2026) & 69.31 & 51.00 & 78.14 & \underline{56.27} & \underline{70.80} & 63.39 & 63.42 & \underline{54.70} & 86.87 \\
RESTORE (ICML2026) & 73.33 & 51.00 & 78.01 & 56.00 & 66.24 & \textbf{64.27} & 78.52 & 51.00 & 88.64 \\
\textbf{Ours} & \textbf{81.53} & \underline{51.44} & \textbf{82.74} & \textbf{59.20} & \textbf{78.24} & \underline{63.77} & \underline{84.15} & \textbf{70.16} & \textbf{96.84} \\
\midrule
\multicolumn{10}{c}{\textit{Prune 80\% of vision tokens}} \\
\midrule
FastV (ECCV2024) & 67.10 & 47.11 & 69.43 & 47.27 & 29.12 & 61.51 & \underline{83.83} & 33.96 & 75.83 \\
SparseVLM (ICML2025) & \underline{70.36} & 50.22 & 74.51 & 52.40 & 33.96 & 62.99 & 72.45 & 40.36 & 79.11 \\
VisPruner (ICCV2025) & 67.88 & 50.22 & 74.48 & 49.73 & 54.40 & 62.03 & 73.52 & 42.00 & \underline{81.49} \\
ZOO-Prune (CVPR2026) & 59.08 & \textbf{51.67} & \underline{74.58} & \underline{53.80} & \underline{59.92} & 62.70 & 53.26 & \underline{43.53} & 79.43 \\
RESTORE (ICML2026) & 64.80 & 50.22 & 74.42 & 51.87 & 53.28 & \textbf{64.18} & 71.00 & 40.35 & 81.05 \\
\textbf{Ours} & \textbf{80.48} & \underline{51.33} & \textbf{81.96} & \textbf{57.47} & \textbf{75.20} & \underline{63.49} & \textbf{83.95} & \textbf{64.57} & \textbf{94.79} \\
\midrule
\multicolumn{10}{c}{\textit{Prune 90\% of vision tokens}} \\
\midrule
FastV (ECCV2024) & 49.64 & 45.89 & 66.52 & 38.40 & 19.40 & 55.95 & \underline{76.82} & 27.99 & 66.16 \\
SparseVLM (ICML2025) & \underline{54.24} & \underline{50.00} & \underline{70.43} & \underline{46.40} & 20.40 & 58.93 & 58.80 & 29.73 & \underline{68.27} \\
VisPruner (ICCV2025) & 45.48 & 48.33 & 68.10 & 42.80 & 30.88 & 59.08 & 53.94 & 30.43 & 66.44 \\
ZOO-Prune (CVPR2026) & 42.04 & 49.67 & 68.01 & 46.33 & \underline{41.00} & 60.71 & 38.17 & \underline{31.75} & 66.56 \\
RESTORE (ICML2026) & 46.70 & 49.56 & 68.39 & 46.00 & 30.76 & \underline{61.57} & 52.27 & 30.26 & 67.79 \\
\textbf{Ours} & \textbf{76.93} & \textbf{50.56} & \textbf{79.86} & \textbf{55.53} & \textbf{69.12} & \textbf{62.88} & \textbf{80.71} & \textbf{56.83} & \textbf{90.65} \\
\bottomrule
\end{tabular}%
}
\endgroup
\end{table}

\begin{table}[t]
\centering
\caption{LLaVA-OneVision-1.5-8B results across eight benchmarks. Avg. Rel.: mean performance relative to Dense (\%); higher is better. Bold/underline: best/second-best pruned results at each ratio.}
\label{tab:main-ov15}
\begingroup
\resizebox{\linewidth}{!}{%
\begin{tabular}{@{}lrrrrrrrrr@{}}
\toprule
Method & TextVQA & MMMU & AI2D & MMStar & ChartQA & NoCaps & TextCaps & InfoVQA & Avg. Rel. \\
\midrule
Dense & 79.71 & 55.78 & 84.62 & 67.33 & 86.64 & 110.40 & 123.13 & 77.41 & 100.00 \\
\midrule
\multicolumn{10}{c}{\textit{Prune 70\% of vision tokens}} \\
\midrule
FastV (ECCV2024) & 66.67 & 54.33 & 75.84 & 51.80 & 48.20 & 103.94 & \underline{114.19} & 43.81 & 80.84 \\
SparseVLM (ICML2025) & \underline{74.68} & 55.00 & 78.21 & 57.33 & 62.76 & 108.44 & 113.21 & 48.83 & 86.94 \\
VisPruner (ICCV2025) & 50.50 & 52.89 & 74.29 & 54.20 & 53.40 & \underline{109.30} & 75.89 & 37.70 & 74.68 \\
ZOO-Prune (CVPR2026) & 71.09 & \underline{55.22} & \textbf{81.83} & \textbf{62.87} & \underline{77.40} & \textbf{110.07} & 107.16 & \underline{54.16} & \underline{90.54} \\
RESTORE (ICML2026) & 63.90 & 52.56 & 75.58 & 54.13 & 44.60 & 103.38 & 104.05 & 34.78 & 77.33 \\
\textbf{Ours} & \textbf{77.56} & \textbf{56.22} & \underline{81.38} & \underline{62.67} & \textbf{83.40} & 107.32 & \textbf{120.73} & \textbf{64.02} & \textbf{95.20} \\
\midrule
\multicolumn{10}{c}{\textit{Prune 80\% of vision tokens}} \\
\midrule
FastV (ECCV2024) & 57.01 & 51.44 & 73.22 & 47.47 & 33.04 & 98.38 & \underline{102.52} & 36.49 & 72.30 \\
SparseVLM (ICML2025) & \underline{67.95} & \textbf{54.56} & 75.97 & 53.40 & 46.72 & 104.13 & 100.64 & 39.87 & 79.21 \\
VisPruner (ICCV2025) & 41.84 & 52.44 & 71.60 & 50.40 & 42.16 & 105.42 & 67.30 & 31.96 & 68.26 \\
ZOO-Prune (CVPR2026) & 65.04 & 53.33 & \underline{79.95} & \underline{60.20} & \underline{69.68} & \textbf{108.73} & 97.84 & \underline{44.10} & \underline{84.56} \\
RESTORE (ICML2026) & 58.11 & 52.78 & 72.22 & 51.33 & 35.32 & 101.90 & 94.82 & 31.02 & 72.41 \\
\textbf{Ours} & \textbf{75.97} & \underline{54.44} & \textbf{80.38} & \textbf{61.13} & \textbf{81.12} & \underline{106.74} & \textbf{116.32} & \textbf{60.89} & \textbf{92.77} \\
\midrule
\multicolumn{10}{c}{\textit{Prune 90\% of vision tokens}} \\
\midrule
FastV (ECCV2024) & 40.80 & 51.78 & 69.04 & 41.60 & 20.56 & 83.65 & 79.75 & 29.52 & 61.22 \\
SparseVLM (ICML2025) & \underline{56.89} & \underline{52.56} & 72.73 & 49.40 & 27.88 & 94.84 & \underline{86.05} & 30.58 & 69.05 \\
VisPruner (ICCV2025) & 30.18 & 49.56 & 69.82 & 44.00 & 31.52 & 95.50 & 53.59 & 28.38 & 59.70 \\
ZOO-Prune (CVPR2026) & 52.86 & 50.44 & \underline{75.23} & \underline{52.93} & \underline{52.48} & \textbf{104.93} & 80.48 & \underline{33.71} & \underline{73.60} \\
RESTORE (ICML2026) & 46.90 & 50.89 & 70.05 & 47.20 & 26.20 & 98.31 & 76.96 & 28.29 & 65.16 \\
\textbf{Ours} & \textbf{71.66} & \textbf{52.89} & \textbf{78.21} & \textbf{57.80} & \textbf{72.72} & \underline{103.93} & \textbf{101.99} & \textbf{52.52} & \textbf{86.47} \\
\bottomrule
\end{tabular}%
}
\endgroup
\end{table}

\begin{figure}[t]
\centering
\includegraphics[width=\linewidth]{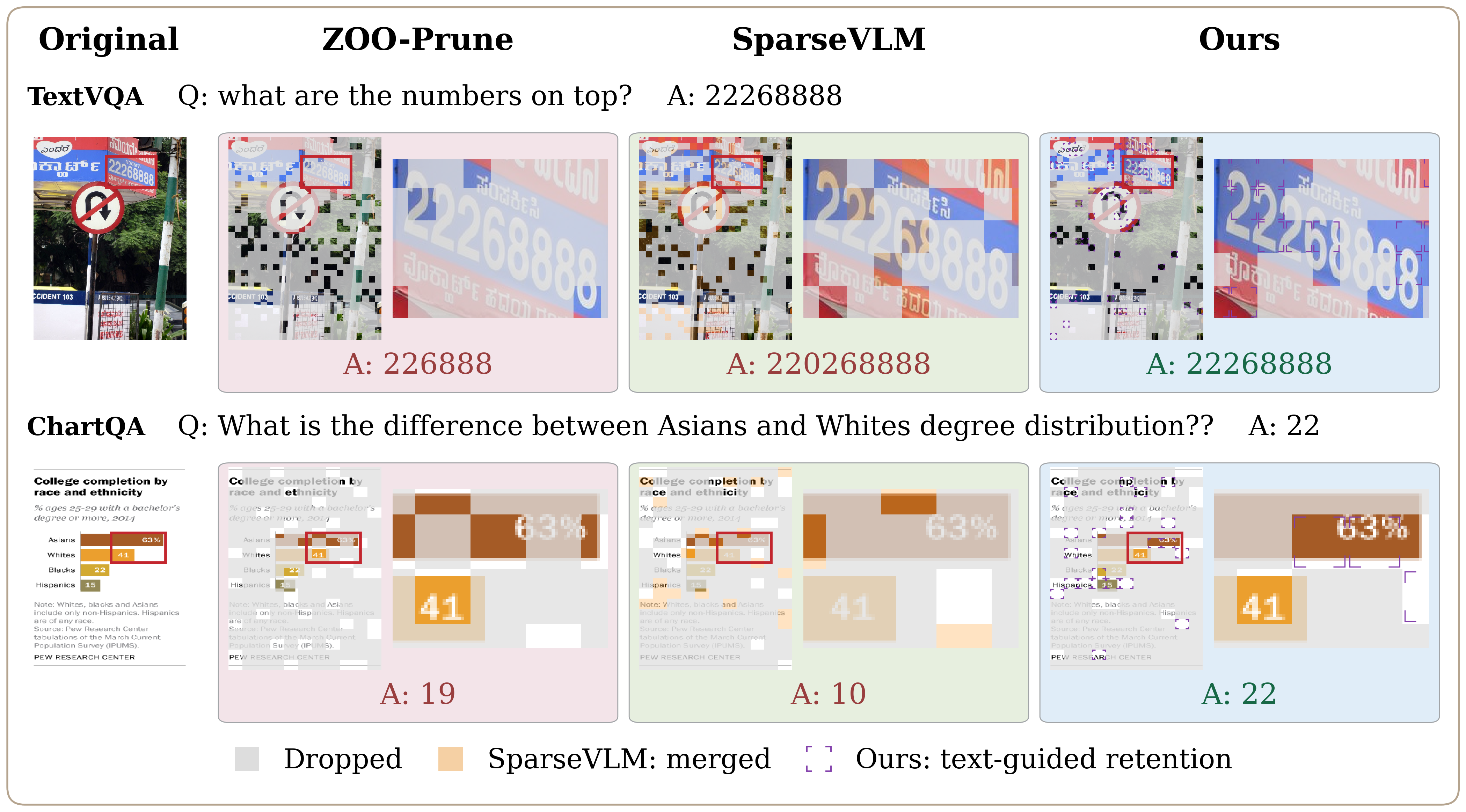}
\caption{TextVQA and ChartQA examples on Qwen3-VL-8B at 80\% pruning. Purple: final tokens outside the image-only top-$K$; gray: discarded regions; orange: merged support.}
\label{fig:qualitative-main}
\end{figure}

\subsection{Main Results}

Tables~\ref{tab:main-qwen3-4b}--\ref{tab:main-ov15} show that our method consistently achieves the highest average relative performance across all nine model--pruning settings. The advantage grows under more aggressive pruning: averaged over the three models, our method exceeds the strongest baseline by 11.10 points at 80\% pruning and 16.84 points at 90\%. Notably, at 90\% pruning on Qwen3-VL-8B, our method retains 90.65\% of Dense performance, while all competing methods remain below 70\%.

The gains are especially pronounced on TextVQA, ChartQA, and InfoVQA, where performance depends on preserving localized, prompt-relevant evidence. Averaged across the three backbones, the margins over the strongest competing results increase from 3.66, 5.51, and 10.91 points at 70\% pruning to 8.16, 11.59, and 15.97 points at 80\%, and further to 19.26, 19.91, and 18.88 points at 90\%, respectively. As pruning becomes more aggressive, early removal is more likely to discard the few tokens containing key evidence; retaining additional candidates until textual guidance becomes informative therefore provides greater benefit in these challenging settings.

Figure~\ref{fig:qualitative-main} further illustrates how deferred selection preserves key evidence. In both examples, answer-relevant regions fall outside the initial image-only top-$K$ but remain in the candidate reserve. Midpoint text-to-visual attention later promotes them into the final set, preserving localized evidence that immediate pruning would discard. This shows how the reserve allows textual guidance to revise the initial visual ranking when relevant evidence is not visually dominant.

\subsection{Ablation Studies}

\begin{figure}[t]
\centering
\includegraphics[width=\linewidth]{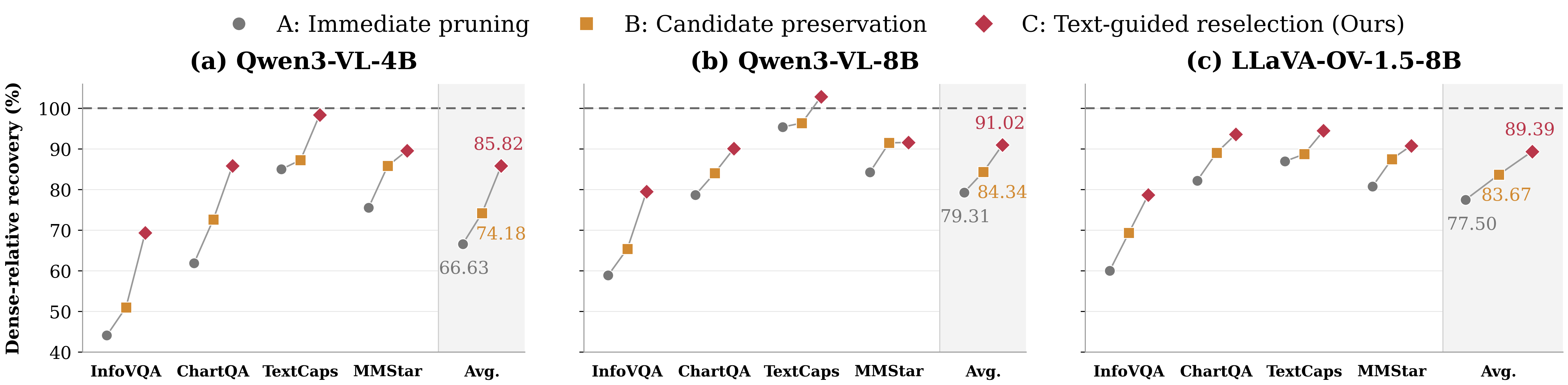}
\caption{
Candidate preservation and text-guided reselection at 80\% pruning.
A: immediate image-only top-$K$ selection;
B: preserve additional candidates until the midpoint, then retain the same initial top-$K$ as A;
C: use the same candidates and token-count trajectory as B, but reselect the final $K$ tokens using text-to-visual attention (Ours). Values are Dense-relative recovery (\%); the dashed line denotes Dense (100\%).
}
\label{fig:ablation_main}
\end{figure}

Figure~\ref{fig:ablation_main} separates the contributions of candidate preservation and text-guided reselection. Immediate image-only pruning retains the image-only top-$K$ from the beginning. Candidate preservation instead keeps the larger candidate pool until the midpoint but ultimately retains the same image-only top-$K$, isolating the benefit of delaying the final pruning decision. Text-guided reselection follows the same token-count trajectory as candidate preservation but reselects the final $K$ tokens using midpoint text-to-visual attention. The substantial gain from candidate preservation to text-guided reselection shows that the improvement is not simply due to retaining more tokens early, but also to using midpoint text-to-visual attention to revise the final selection.

\begin{figure}[t]
\centering
\includegraphics[width=\linewidth]{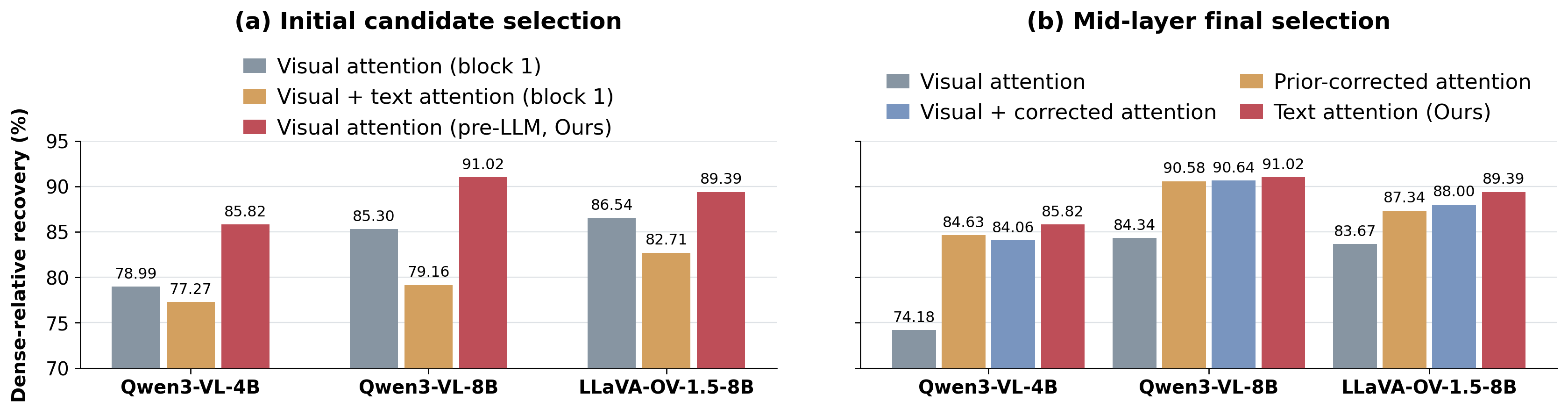}
\caption{
Pruning-score ablations at 80\% pruning.
(a) Encoder-based candidate selection before the LLM versus after block 1, with or without text attention.
(b) Midpoint scoring with fixed candidates and token budgets.
Results average Dense-relative recovery across InfoVQA, ChartQA, TextCaps, and MMStar.
}
\label{fig:stage-scoring-ablation}
\end{figure}

Figure~\ref{fig:stage-scoring-ablation} examines which information should guide token selection at each stage. In Figure~\ref{fig:stage-scoring-ablation}(a), using text-to-visual attention in the first decoder block reduces recovery compared with vision-only pre-LLM pruning across all three models, showing that textual guidance can be detrimental when applied too early. In contrast, Figure~\ref{fig:stage-scoring-ablation}(b) shows that, with the candidate pool, reselection depth, and final budget fixed, midpoint text-to-visual attention alone achieves the highest average recovery on all three models. These results support a clear separation of roles: vision-encoder attention for early pruning and text-to-visual attention for midpoint reselection.

\subsection{Analysis}

\begin{figure}[t]
\centering
\IfFileExists{figures/alpha_prefill_performance.png}{%
  \includegraphics[width=\linewidth]{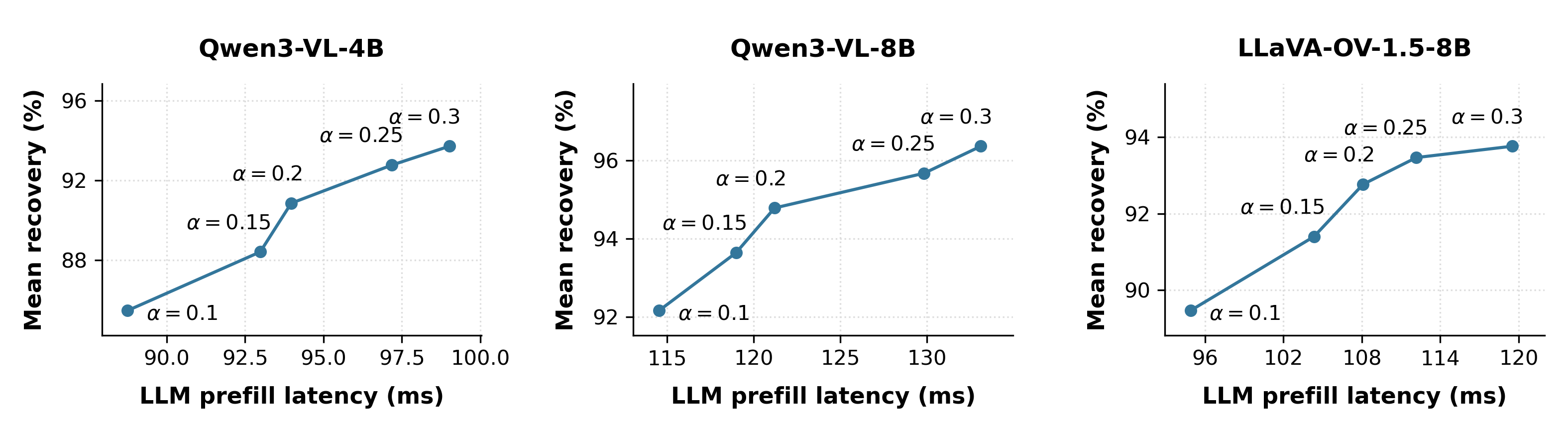}%
}{}
\caption{
Performance--latency trade-off as the reserve fraction $\alpha$ varies at 80\% pruning.
Recovery is averaged equally across eight benchmarks relative to Dense.
Latency includes token-selection overhead and is averaged over three timed repetitions on one RTX A6000 GPU, using 100 fixed InfoVQA inputs stratified by image size and aspect ratio.
}
\label{fig:alpha-flops}
\end{figure}

The reserve fraction $\alpha$ controls the trade-off between candidates preserved for midpoint reselection and inference cost. Figure~\ref{fig:alpha-flops} shows that increasing $\alpha$ improves average recovery across all three models at the cost of higher LLM-prefill latency. Across $\alpha\in[0.1,0.3]$, this trade-off is approximately linear, allowing $\alpha$ to be adjusted to the desired efficiency--accuracy balance. Even $\alpha=0.1$ exceeds the strongest baseline average on every backbone, and we use $\alpha=0.2$ by default.

Table~\ref{tab:efficiency} shows that our method maintains practical inference latency despite preserving additional visual candidates in the early decoder layers. Unlike many existing approaches that introduce additional token-merging or learned selection modules, our method simply uses vision-encoder attention for early pruning and text-to-visual attention for midpoint reselection. This targeted use of existing model signals yields substantial performance gains with only modest additional inference cost. Additional analyses in the Appendix quantify the use and performance contribution of reselected candidates and extend the comparison to additional progressive pruning methods.

\begin{table}[t]
\centering
\caption{
Qwen3-VL-8B inference latency (ms).
Measurements average three timed repetitions on one RTX A6000 GPU using 100 fixed InfoVQA inputs stratified by image size and aspect ratio.
Prefill includes token-selection overhead.
End-to-end spans GPU-ready inputs through the first output token, excluding CPU preprocessing and input transfer.
}
\label{tab:efficiency}
\begin{tabular}{lrrrr}
\toprule
& \multicolumn{2}{c}{80\% pruning} & \multicolumn{2}{c}{90\% pruning} \\
\cmidrule(lr){2-3}\cmidrule(lr){4-5}
Method & LLM prefill & End-to-end & LLM prefill & End-to-end \\
\midrule
Dense & 226.66 & 364.13 & 226.66 & 364.13 \\
FastV & 116.46 & 256.15 & 106.97 & 247.22 \\
SparseVLM & 134.79 & 276.36 & 122.16 & 263.63 \\
VisPruner & 131.25 & 292.39 & 123.80 & 284.79 \\
ZOO-Prune & 137.28 & 401.81 & 111.69 & 375.04 \\
RESTORE & 128.87 & 288.37 & 116.46 & 276.09 \\
Ours & 121.20 & 282.35 & 108.37 & 268.50 \\
\bottomrule
\end{tabular}
\end{table}

\section{Conclusion}

We show that textual guidance is most effective for visual token pruning when applied at an appropriate decoder depth. Based on this observation, we introduce a simple training-free method that uses vision-encoder attention for early pruning and text-to-visual attention for midpoint reselection, achieving strong performance preservation across three VLMs and eight benchmarks, especially under aggressive pruning. Despite its simplicity, the method provides substantial gains with practical inference efficiency. We hope this approach can serve as a strong baseline for future work toward more effective and general visual token pruning in large vision-language models.

\bibliography{iclr2027_conference}
\bibliographystyle{iclr2027_conference}
\appendix

\section{Additional Protocol Details}
\label{app:protocol}

\paragraph{Benchmarks and metrics.}
The full evaluation sets contain 5,000 TextVQA \citep{textvqa}, 900 MMMU \citep{mmmu}, 3,088 AI2D \citep{ai2d}, 1,500 MMStar \citep{mmstar}, 2,500 ChartQA \citep{chartqa}, 4,500 NoCaps \citep{nocaps}, 3,166 TextCaps \citep{textcaps}, and 2,801 InfoVQA \citep{infovqa} examples. TextVQA uses VQA scoring without additional OCR text in the prompt; MMMU, AI2D, and MMStar use accuracy; ChartQA uses relaxed correctness; NoCaps and TextCaps use CIDEr; and InfoVQA uses ANLS. CIDEr and ANLS are displayed after multiplication by 100. \emph{Avg. Rel.} is the equal-weighted mean of each task score normalized by the corresponding Dense score and can exceed 100\%. Four-task pruning ablations average InfoVQA, ChartQA, TextCaps, and MMStar equally, whereas the depth diagnostic in Figure~\ref{fig:question-depth-design} uses TextVQA, ChartQA, InfoVQA, and AI2D.

\paragraph{Evaluation and baseline adaptation.}
For task evaluation, Qwen models use FlashAttention-2 except for RESTORE, whose implementation requires SDPA to apply
additive attention biases. All methods on LLaVA-OneVision-1.5 use SDPA. KV caching is disabled for all methods to maintain a consistent. evaluation configuration. Within each backbone, all methods share inputs, preprocessing, generation settings, and scorers. Baselines are adapted from their official implementations. FastV retains two initial decoder blocks before pruning. Since SparseVLM progressively changes the number of visual tokens across decoder depth, we match its average decoder visual-token usage to ours rather than its final token count. VisPruner, ZOO-Prune, and RESTORE perform their initial visual-token reduction before the decoder; RESTORE additionally applies position and attention correction inside the language model. Appendix~\ref{app:progressive-comparison} provides additional comparisons with progressive pruning methods.

\subsection{Selection Implementation}
\label{app:implementation}

\paragraph{Encoder scores.}
Qwen aggregates vision-encoder attention received by each visual key across visual queries and heads. LLaVA-OneVision uses head-averaged CLS-to-patch attention from the penultimate vision-encoder block. Scores are aligned with each backbone's visual-token geometry before top-$M$ selection.

\paragraph{Midpoint scores.}
Midpoint scores are computed from the decoder queries and keys using the input prompt tokens as queries. Correct answers and generated answer tokens are not used for token selection. When attention probabilities are not directly exposed by the attention kernel, the required scores are recomputed from the corresponding queries and keys.

\paragraph{Computational budget.}
For reselection after block $L$, we measure visual-token processing using the token--block count
\begin{equation}
B_{\mathrm{vis}}
=
LM+(D-L)K
\approx
DK+\alpha L(N-K).
\label{eq:token-blocks}
\end{equation}
This measure is used to match decoder visual-token workloads across selection depths and progressive pruning baselines. For the analytical FLOPs results in Figure~\ref{fig:alpha-appendix-tradeoffs}, decoder cost is computed from the sequence length processed before and after reselection and averaged over individual inputs.

\subsection{Reserve-Size Trade-off}
\label{app:reserve-tradeoff}

\begin{figure}[t]
\centering
\includegraphics[
    width=\linewidth
]{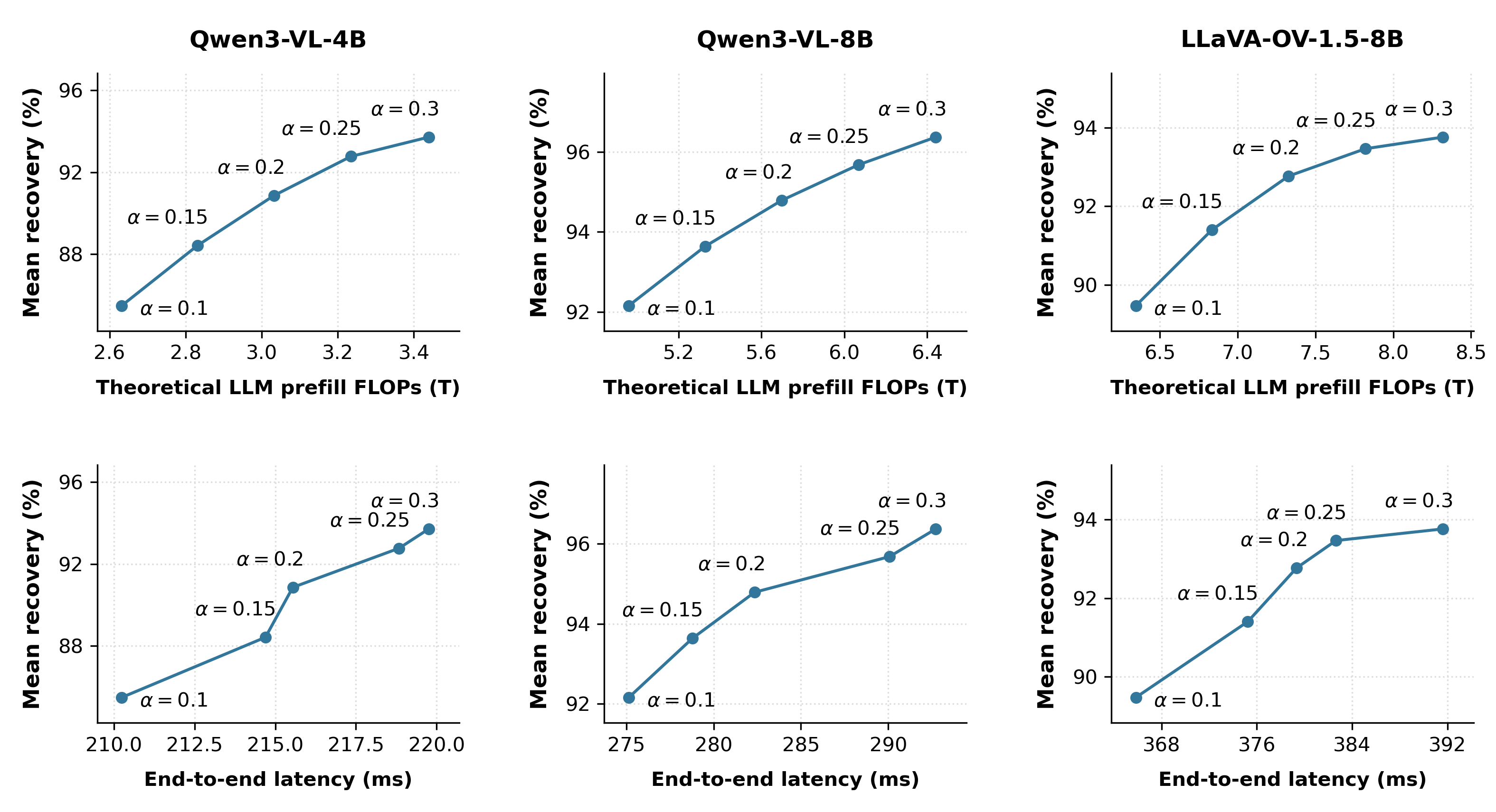}
\caption{
Reserve-size trade-offs at 80\% pruning across the three backbones.
Recovery is averaged equally across eight benchmarks relative to Dense.
Top: analytical decoder-prefill FLOPs, excluding the vision frontend and token-selection overhead.
Bottom: measured GPU-input-to-first-token latency, including the vision frontend and token-selection overhead but excluding CPU preprocessing and input transfer.
Labels indicate $\alpha\in\{0.10,0.15,0.20,0.25,0.30\}$, the fraction of otherwise pruned tokens preserved as additional candidates until midpoint reselection.
}
\label{fig:alpha-appendix-tradeoffs}
\end{figure}

Figure~\ref{fig:alpha-appendix-tradeoffs} examines the effect of the reserve fraction $\alpha$ using two measures of computational cost: analytical decoder-prefill FLOPs and GPU-input-to-first-token latency. Each point corresponds to $\alpha\in\{0.10,0.15,0.20,0.25,0.30\}$ at 80\% pruning, with recovery averaged equally over all eight benchmarks.

Across all three backbones, larger reserves improve recovery while increasing computation, with diminishing gains at larger reserve fractions. The same trend under analytical FLOPs confirms that this trade-off is not specific to measured latency. The relative increase in first-token latency is smaller because it includes vision processing and other reserve-independent costs. Overall, $\alpha$ directly controls the trade-off between candidate preservation and inference cost.

\section{Additional Ablation Analysis}
\label{app:additional-ablation}

\paragraph{Effect of candidate preservation and reselection.}
Table~\ref{tab:reserve-reselection} reports the fraction of final tokens selected from outside the initial vision-guided top-$K$, together with the recovery gain from candidate preservation to text-guided reselection. Across tasks and backbones, approximately 39--45\% of the final visual tokens come from the reserved candidates, showing that midpoint text-to-visual attention substantially revises the initial visual ranking. The largest recovery gains appear on InfoVQA and ChartQA.

\begin{table}[t]
\centering
\caption{
Effect of candidate preservation and midpoint reselection at 80\% pruning.
Ret. is the input-averaged percentage of final tokens selected from outside the initial vision-guided top-$K$.
$\Delta$ is the Dense-relative recovery gain (percentage points) from candidate preservation to text-guided reselection under the same token-count trajectory.
}
\label{tab:reserve-reselection}
\begin{tabular}{lrrrrrr}
\toprule
& \multicolumn{2}{c}{Qwen3-VL-4B}
& \multicolumn{2}{c}{Qwen3-VL-8B}
& \multicolumn{2}{c}{LLaVA-OV-1.5-8B} \\
\cmidrule(lr){2-3}\cmidrule(lr){4-5}\cmidrule(lr){6-7}
Task
& Ret. (\%) & $\Delta$ (pp)
& Ret. (\%) & $\Delta$ (pp)
& Ret. (\%) & $\Delta$ (pp) \\
\midrule
InfoVQA
& 42.09 & $+18.37$
& 42.96 & $+14.09$
& 44.28 & $+9.28$ \\
ChartQA
& 39.03 & $+13.27$
& 42.00 & $+6.04$
& 41.18 & $+4.57$ \\
TextCaps
& 40.65 & $+11.15$
& 41.47 & $+6.47$
& 44.74 & $+5.75$ \\
MMStar
& 43.84 & $+3.78$
& 41.55 & $+0.11$
& 44.38 & $+3.27$ \\
\bottomrule
\end{tabular}
\end{table}

\section{Comparison with Progressive Pruning}
\label{app:progressive-comparison}

Table~\ref{tab:progressive-comparison} extends our evaluation to two additional progressive pruning methods, PyramidDrop \citep{xing2024pyramiddrop} and FlowCut \citep{flowcut}. To account for their different pruning schedules, we match each baseline to Ours' per-input visual token--block budget at $\alpha=0.2$, up to integer rounding. The 80\% and 90\% settings denote Ours' final pruning ratios; progressive baselines may therefore use different final token counts under the matched workloads. Ours achieves the highest average relative performance at both budgets. At 80\% pruning, Ours exceeds FlowCut and PyramidDrop by 9.51 and 19.26 percentage points, respectively; at 90\%, the margins increase to 9.47 and 27.82 points. 

\begin{table}[t]
\centering
\caption{
Comparison with two additional progressive pruning methods on Qwen3-VL-8B under matched per-input visual token--block budgets.
The 80\% and 90\% settings denote Ours' final pruning ratios; baseline final token counts may differ under the matched workloads.
Avg. Rel. is the eight-task mean performance relative to Dense.
Bold/underline indicate the best/second-best pruned results.
}
\label{tab:progressive-comparison}
\begingroup
\resizebox{\linewidth}{!}{%
\begin{tabular}{@{}lrrrrrrrrr@{}}
\toprule
Method & TextVQA & MMMU & AI2D & MMStar & ChartQA & NoCaps & TextCaps & InfoVQA & Avg. Rel. \\
\midrule
Dense
& 82.95 & 51.67 & 83.68 & 62.73 & 83.48 & 63.54 & 81.62 & 81.19 & 100.00 \\
\midrule
\multicolumn{10}{c}{\textit{Matched to Ours at 80\% pruning}} \\
\midrule
PyramidDrop
& 61.77 & 48.22 & 70.76 & 47.33 & 45.80 & 62.41 & 66.92 & 33.58 & 75.53 \\
FlowCut
& \underline{77.23} & \underline{50.44} & \underline{75.71} & \underline{51.07}
& \underline{50.40} & \underline{62.47} & \underline{83.83} & \underline{47.29}
& \underline{85.28} \\
\textbf{Ours}
& \textbf{80.48} & \textbf{51.33} & \textbf{81.96} & \textbf{57.47}
& \textbf{75.20} & \textbf{63.49} & \textbf{83.95} & \textbf{64.57}
& \textbf{94.79} \\
\midrule
\multicolumn{10}{c}{\textit{Matched to Ours at 90\% pruning}} \\
\midrule
PyramidDrop
& 43.18 & 47.78 & 67.78 & 41.33 & 25.28 & 57.12 & 48.34 & 25.87 & 62.83 \\
FlowCut
& \underline{72.50} & \underline{49.89} & \underline{71.99} & \underline{49.87}
& \underline{44.32} & \underline{61.64} & \underline{79.10} & \underline{42.97}
& \underline{81.18} \\
\textbf{Ours}
& \textbf{76.93} & \textbf{50.56} & \textbf{79.86} & \textbf{55.53}
& \textbf{69.12} & \textbf{62.88} & \textbf{80.71} & \textbf{56.83}
& \textbf{90.65} \\
\bottomrule
\end{tabular}%
}
\endgroup
\end{table}

\section{Qualitative Results}
\label{app:qualitative-results}

\begin{figure}[p]
\centering
\includegraphics[width=\linewidth]{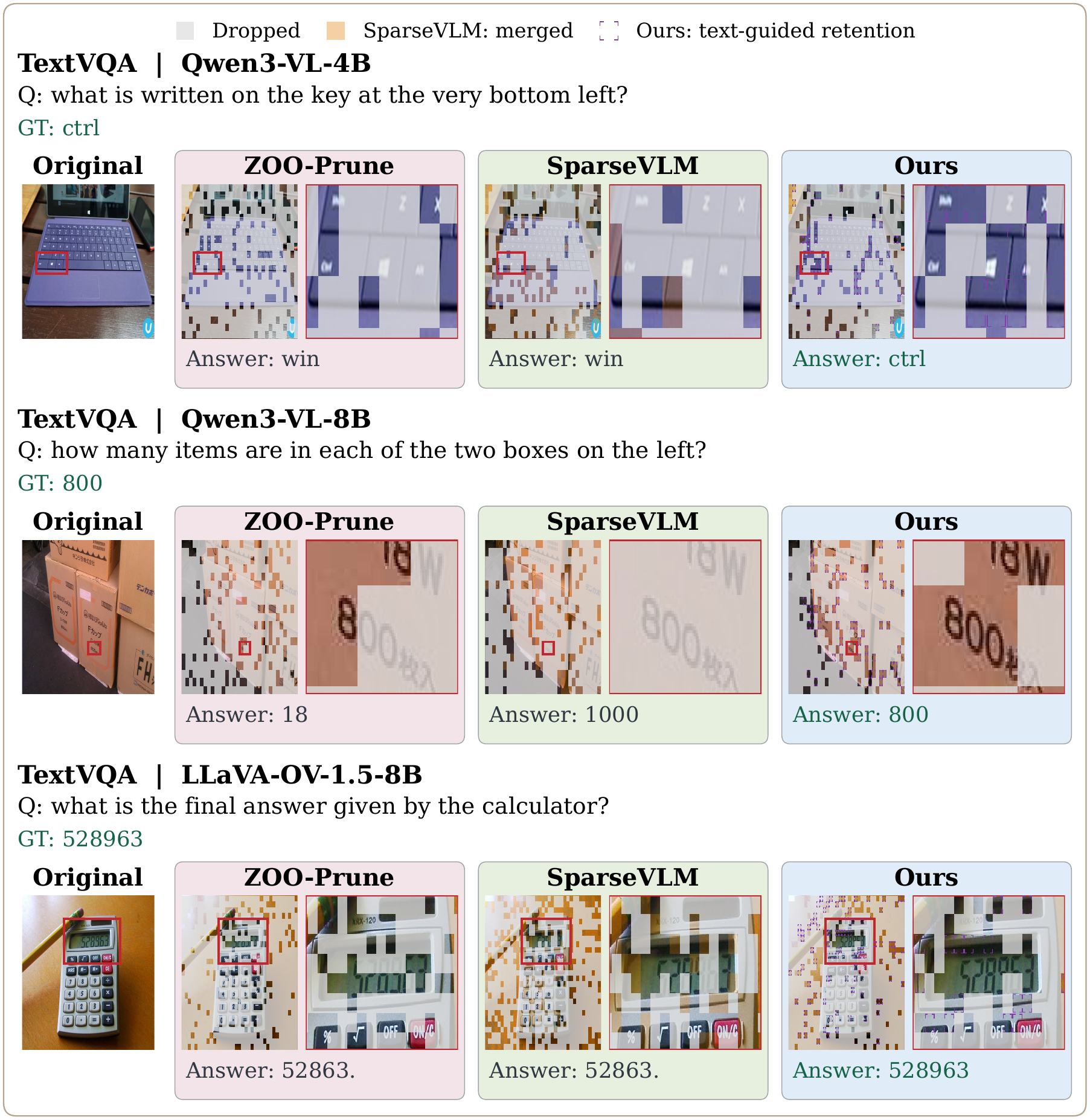}
\caption{
TextVQA examples across three backbones at 80\% pruning.
Columns compare the original image, ZOO-Prune, SparseVLM, and Ours.
}
\label{fig:qualitative-textvqa}
\end{figure}

\begin{figure}[p]
\centering
\includegraphics[width=\linewidth]{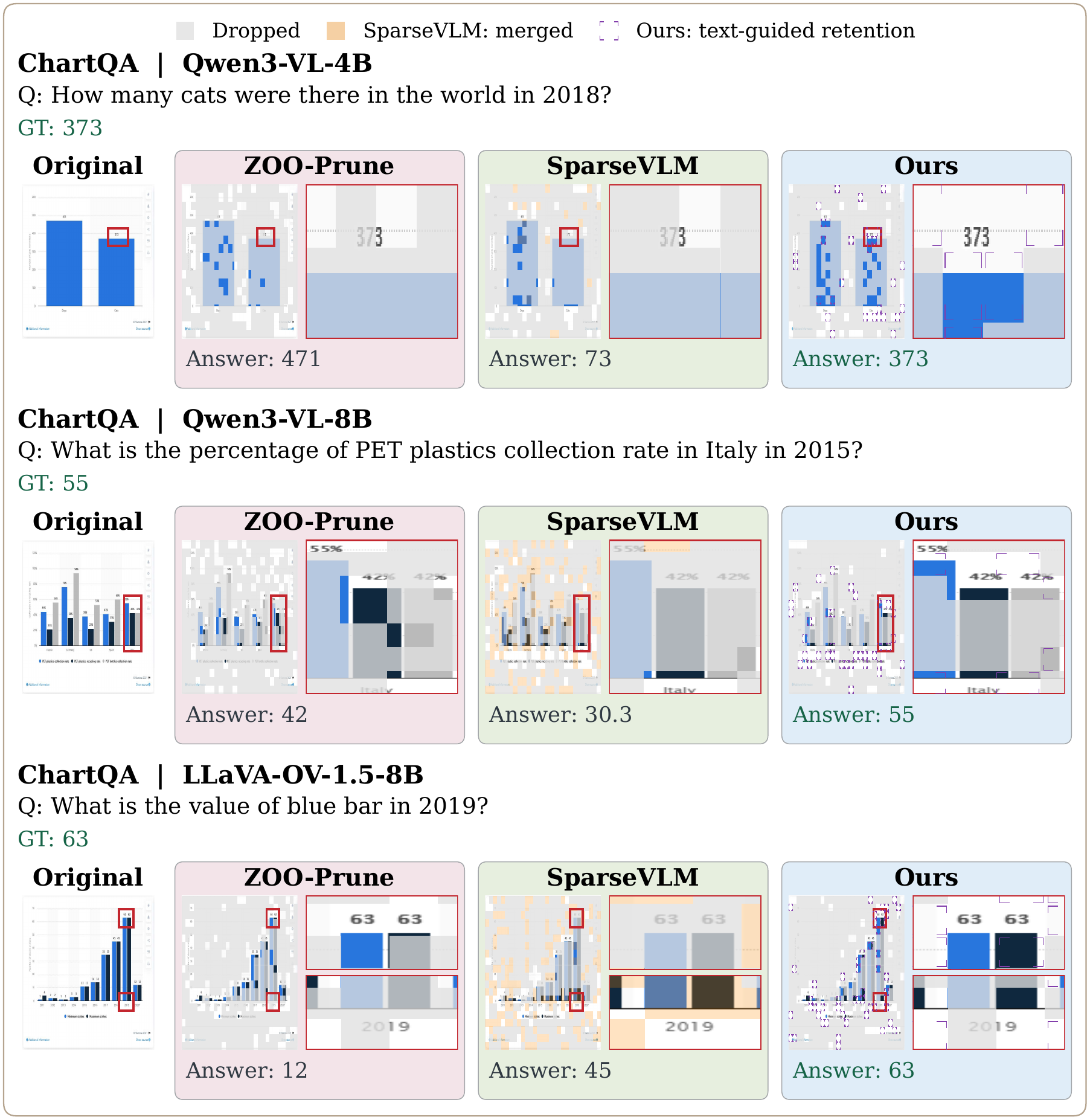}
\caption{
ChartQA examples across three backbones at 80\% pruning.
Columns compare the original image, ZOO-Prune, SparseVLM, and Ours.
}
\label{fig:qualitative-chartqa}
\end{figure}

\begin{figure}[p]
\centering
\includegraphics[width=\linewidth]{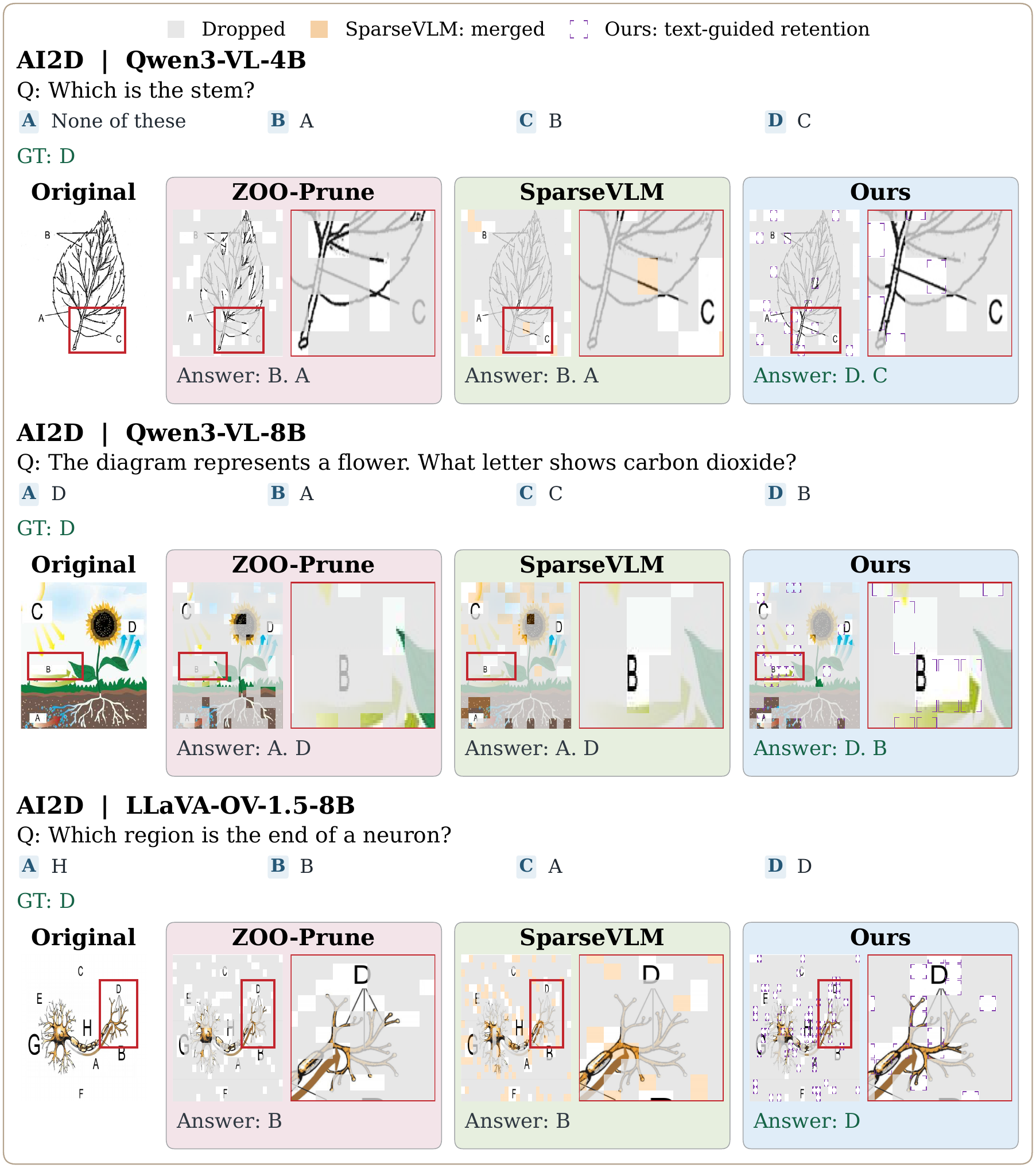}
\caption{
AI2D examples across three backbones at 80\% pruning.
Columns compare the original image, ZOO-Prune, SparseVLM, and Ours.
}
\label{fig:qualitative-ai2d}
\end{figure}

\begin{figure}[p]
\centering
\includegraphics[width=\linewidth]{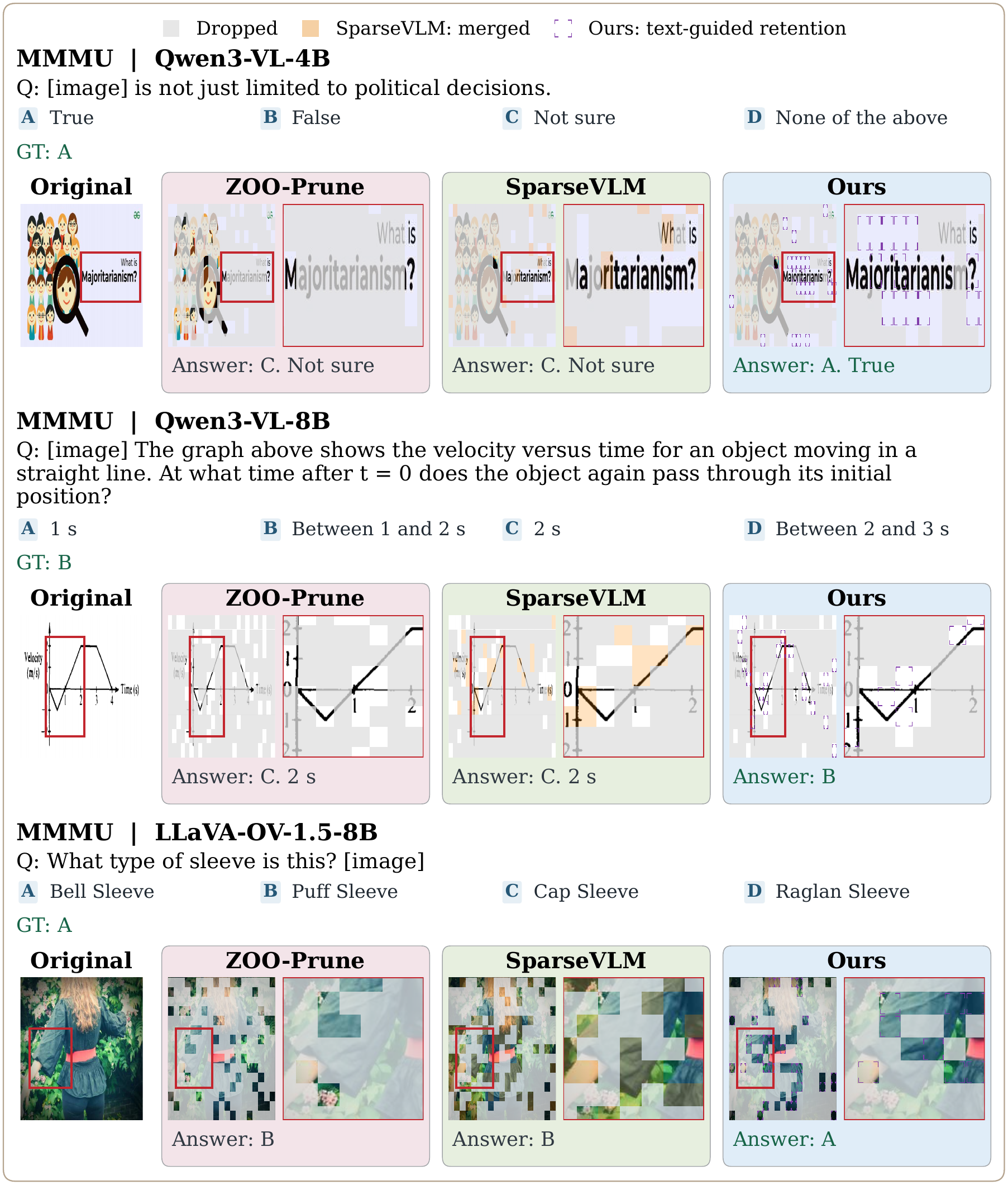}
\caption{
MMMU examples across three backbones at 80\% pruning.
Columns compare the original image, ZOO-Prune, SparseVLM, and Ours.
}
\label{fig:qualitative-mmmu}
\end{figure}

\begin{figure}[p]
\centering
\includegraphics[width=\linewidth]{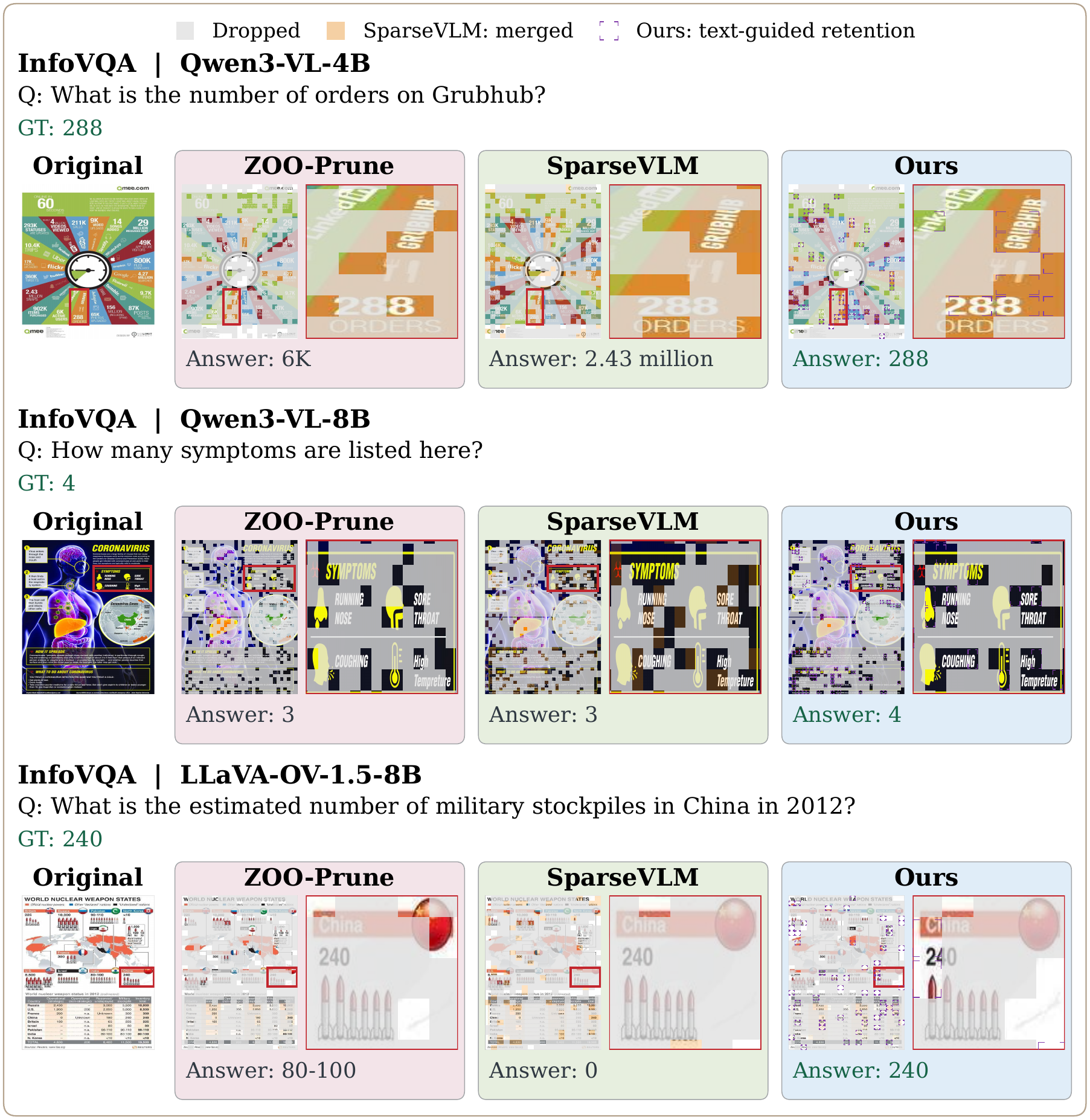}
\caption{
InfoVQA examples across three backbones at 80\% pruning.
Columns compare the original image, ZOO-Prune, SparseVLM, and Ours.
}
\label{fig:qualitative-infovqa}
\end{figure}

\begin{figure}[p]
\centering
\includegraphics[width=\linewidth]{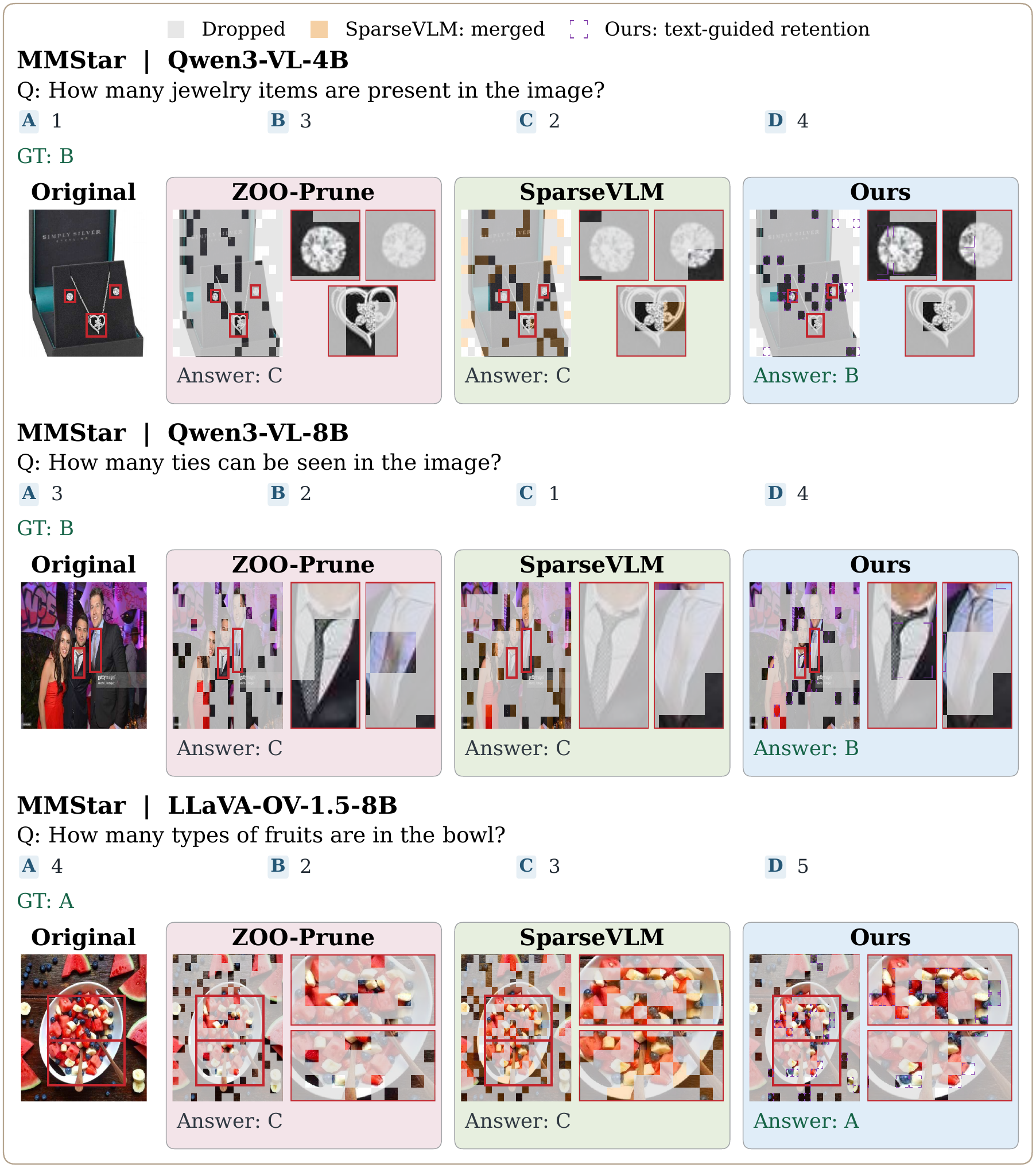}
\caption{
MMStar examples across three backbones at 80\% pruning.
Columns compare the original image, ZOO-Prune, SparseVLM, and Ours.
}
\label{fig:qualitative-mmstar}
\end{figure}

\begin{figure}[p]
\centering
\includegraphics[width=\linewidth]{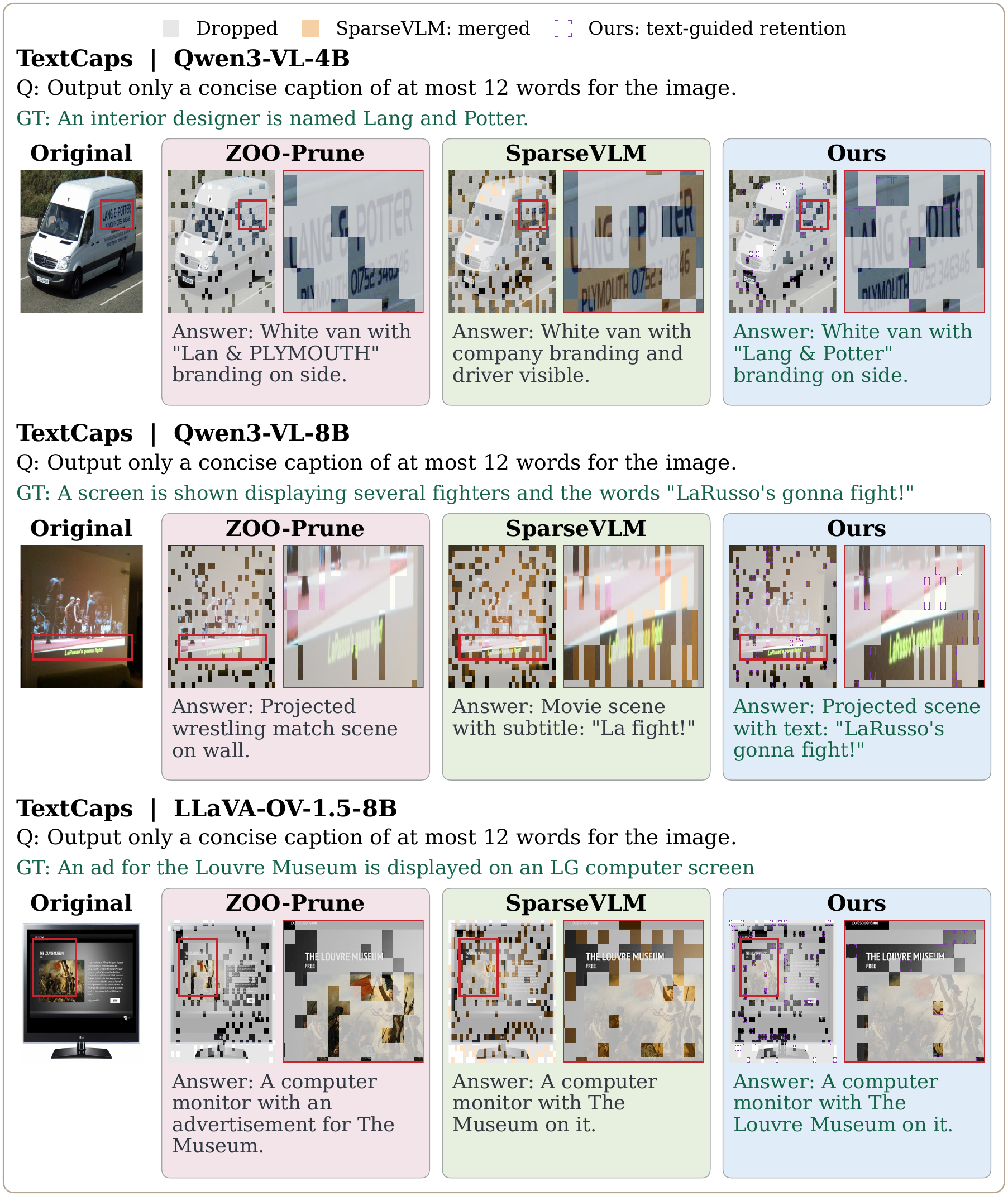}
\caption{
TextCaps examples across three backbones at 80\% pruning.
Columns compare the original image, ZOO-Prune, SparseVLM, and Ours.
}
\label{fig:qualitative-textcaps}
\end{figure}

\begin{figure}[p]
\centering
\includegraphics[width=\linewidth]{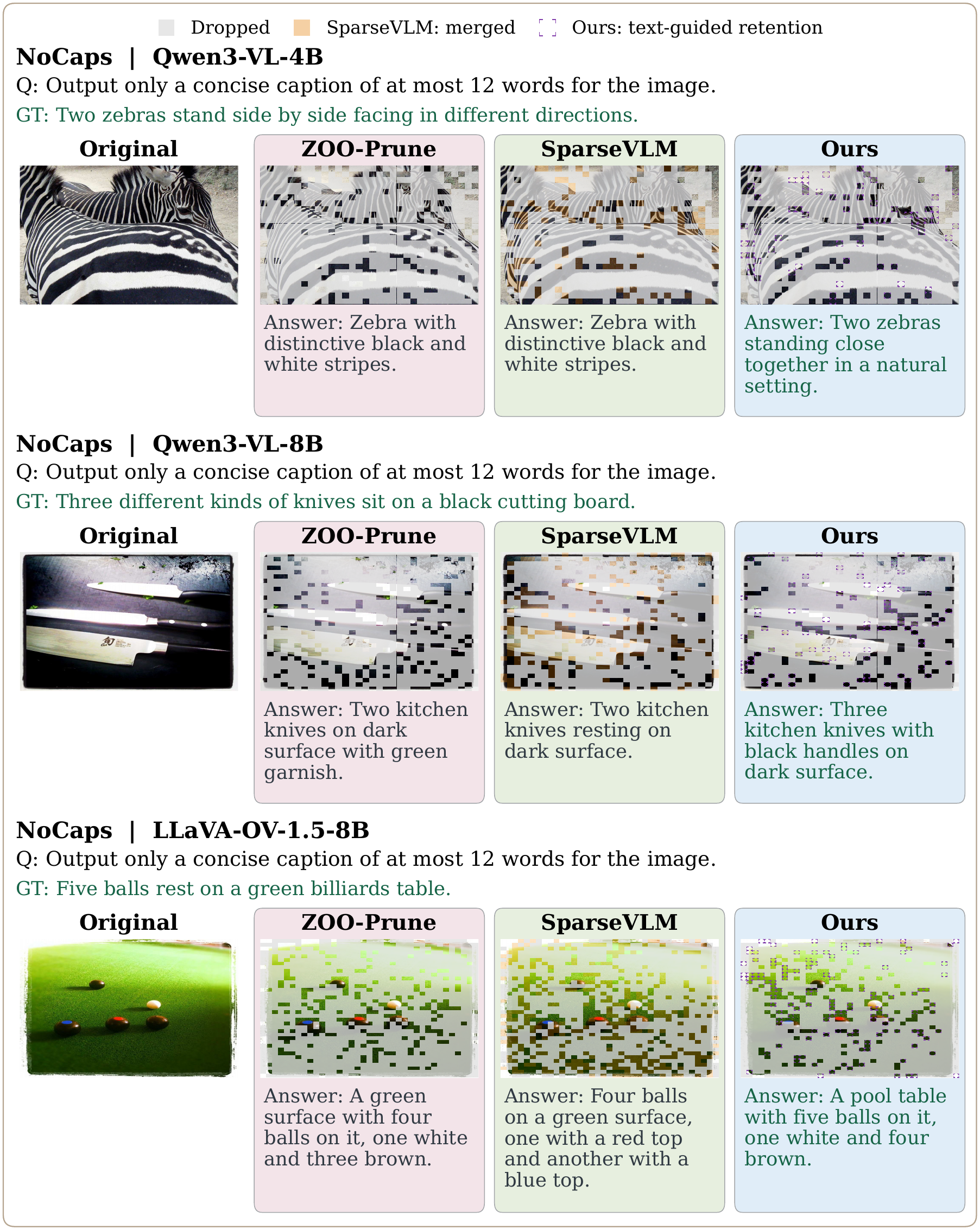}
\caption{
NoCaps examples across three backbones at 80\% pruning.
Columns compare the original image, ZOO-Prune, SparseVLM, and Ours.
}
\label{fig:qualitative-nocaps}
\end{figure}

Figures~\ref{fig:qualitative-textvqa}--\ref{fig:qualitative-nocaps} provide qualitative comparisons across all eight benchmarks at 80\% pruning on three VLMs. The examples highlight a recurring limitation of existing pruning strategies: visually small but task-relevant evidence can be removed even when surrounding or semantically related regions are preserved. This is particularly visible on TextVQA, ChartQA, and InfoVQA, where answering correctly often depends on retaining localized text, numbers, or chart elements. ZOO-Prune can preserve visually salient regions while missing such details, whereas SparseVLM can retain related visual support yet lose the precise evidence or relationships required for the answer.

In contrast, our deferred reselection can preserve these initially lower-ranked candidates until text-to-visual attention becomes more informative, allowing the final token set to better retain question-relevant evidence. The AI2D, MMMU, and MMStar examples further show that this behavior extends beyond scene text to diagrams and general multimodal reasoning, where relevant evidence may be spatially sparse or distributed across multiple regions. Finally, the TextCaps and NoCaps examples indicate that the benefit is not limited to question-conditioned tasks: our method also preserves visual information needed for more faithful open-ended descriptions. Together, these examples qualitatively support the main finding that delaying text-guided selection helps preserve task-relevant evidence that early pruning may otherwise discard.

\end{document}